%% file: acl_latex.tex
\pdfoutput=1

\documentclass[11pt]{article}

\usepackage{acl}

\usepackage{times}
\usepackage{latexsym}

\usepackage[T1]{fontenc}

\usepackage[utf8]{inputenc}

\usepackage{microtype}

\usepackage{inconsolata}
\usepackage{graphicx}
\usepackage{multirow, array}
\usepackage{amssymb}
\usepackage{pifont}
\usepackage{booktabs}
\usepackage{graphicx}
\usepackage{multirow}
\usepackage{url}
\usepackage{subcaption}
\usepackage{hyperref}
\usepackage{xspace}
\usepackage{makecell}
\usepackage{fontawesome5}
\usepackage{comment}
\usepackage{float}
\usepackage{enumitem}
\title{
\textit{Is the Pope Catholic? Yes, the Pope is Catholic. }\\Generative Evaluation of Non-Literal Intent Resolution in LLMs
}

\newcommand{\aspace}{\hspace{2em}}
\newcommand{\cmu}{$^\heartsuit$}
\newcommand{\aitwo}{$^\clubsuit$}

\author{
Akhila Yerukola\cmu \aspace Saujas Vaduguru\cmu \aspace Daniel Fried\cmu \aspace Maarten Sap\cmu\aitwo\\
\cmu Language Technologies Institute, Carnegie Mellon University \\
\aitwo Allen Institute for AI\\
\faEnvelope~\texttt{\href{mailto:ayerukol@andrew.cmu.edu}{ayerukol@andrew.cmu.edu}}
}

\begin{document}
\maketitle
\input{sections/0-abstract}
\input{sections/1-introduction}

\input{sections/3-response_generation}

\input{sections/4-experiments_for_response}

\input{sections/7-expanding}

\input{sections/2-relatedwork}

\input{sections/8-conclusion}
\input{sections/9-limitations}
\input{sections/10-acknowledgement}

\bibliography{anthology,custom}
\bibliographystyle{acl_natbib}
\pagebreak
\appendix
\input{sections/appendix}

\end{document}

%% file: sections/0-abstract.tex
\begin{abstract}
Humans often express their communicative intents indirectly or non-literally, which requires their interlocutors---human or AI---to understand beyond the literal meaning of words. While most existing work has focused on discriminative evaluations, we present a new approach to generatively evaluate large language models' (LLMs') intention understanding by examining their responses to non-literal utterances. Ideally, an LLM should respond in line with the true intention of a non-literal utterance, not its literal interpretation. Our findings show that LLMs struggle to generate pragmatically relevant responses to non-literal language, achieving only 50-55\% accuracy on average. While explicitly providing oracle intentions significantly improves performance (e.g., 75\% for \texttt{Mistral-Instruct}), this still indicates challenges in leveraging given intentions to produce appropriate responses. Using chain-of-thought to make models spell out intentions yields much smaller gains (60\% for \texttt{Mistral-Instruct}). These findings suggest that LLMs are not yet effective pragmatic interlocutors, highlighting the need for better approaches for modeling intentions \textit{and} utilizing them for pragmatic generation.\footnote{Code and data are available at: \url{https://github.com/Akhila-Yerukola/generative-intention-resolution}.}
\end{abstract}


%% file: sections/1-introduction.tex
\section{Introduction}
Humans possess the ability to communicate and understand each other even through non-literal utterances and conversational implicatures \cite{roberts1994people, dews1999obligatory, glucksberg2001understanding}. This is attributed to their ability to make \textit{pragmatic} inferences arising from contextual factors and conventions in conversation, rather than specific words or phrases \cite{grice1975logic, davis2016implicature}. Since humans often use non-literal language in communication, large language models (LLMs) must also develop pragmatic understanding to facilitate effective and nuanced human-AI interactions.

\begin{figure*}
    \centering
    \includegraphics[trim={0 2em 0 2em}, scale=0.35]{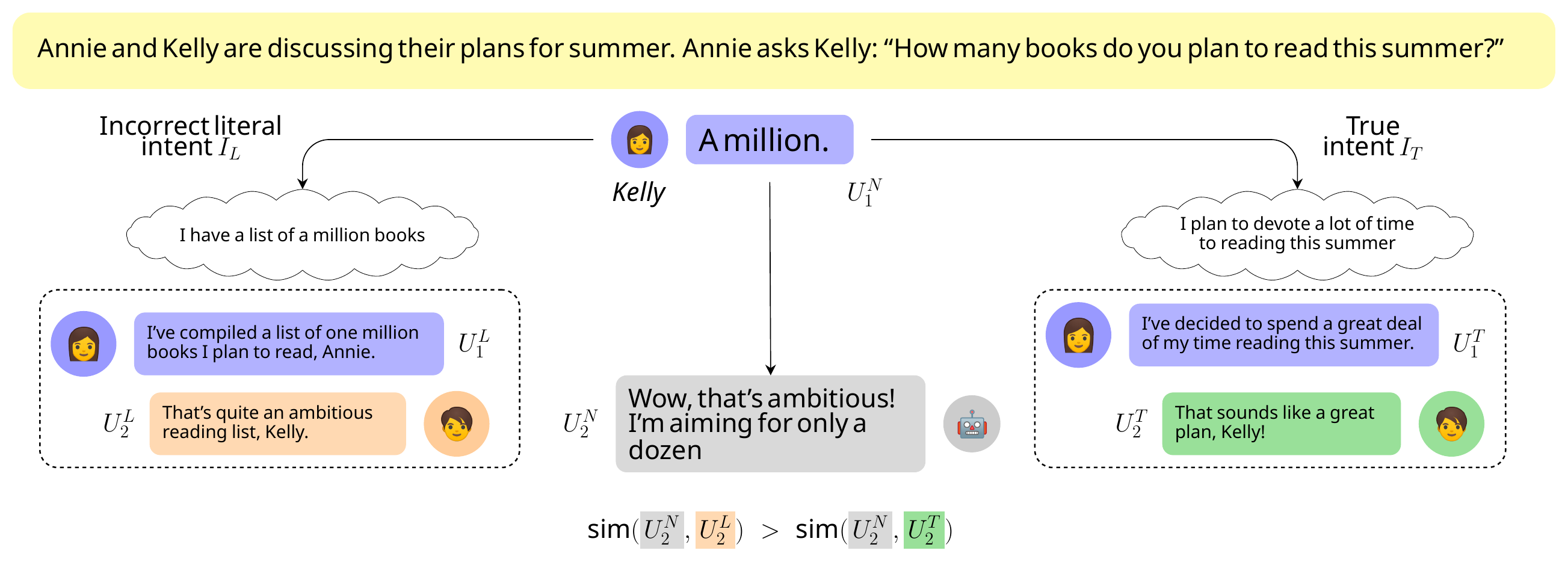}
    \caption{Framework to evaluate whether an LLM can generate an appropriate response to non-literal language use. Given a context $C$ and a non-literal utterance $U_1^N$, the model responds with $U_2^N$. Our proposed framework compares $U_2^N$ against responses ($U_2^L$ and $U_2^T$) from two counterfactual dialog chains based on conveying incorrect literal meaning $I_L$ and direct true intent $I_T$.  We then compare the similarity of the model generated response $U_2^N$ to these reference responses, under the context $C$, to determine whether it is appropriate.}
    \label{fig:figure1}
\end{figure*}

In this work, we introduce a new generative evaluation framework designed to evaluate the ability of LLMs to understand and resolve intentions through pragmatic response generation. In Figure \ref{fig:figure1}, Kelly uses hyperbole to express her desire to read numerous books. A contextually appropriate response would be to ideally echo sentiments like ``That sounds like a great plan'' rather than interpreting ``a million'' literally, as seen in responses like ``That's quite an ambitious reading list''. Our framework uses this intuition to compare LLMs' responses to human-like expectations, enabling a nuanced assessment of their pragmatic understanding and response accuracy.

Our primary focus on \textit{pragmatic response generation} marks a departure from prior work \cite{zheng2021grice,hu2022fine,srivastava2023beyond,ruis2023goldilocks}, which has predominantly measured intention understanding through a \textit{discriminative} contrastive multiple-choice classification. We show that this setting does not necessarily reflect LLMs' abilities in generating pragmatic responses, nor does it correspond to the
use of LLMs as conversational agents \cite{west2023generative}. 

We evaluate the pragmatic understanding of several state-of-the-art open-source LLMs on various types of non-literal language from \citet{hu2022fine}. We observe that LLMs often struggle with generating contextually appropriate responses and tend to interpret non-literal language literally, with an accuracy of 50-55\%.
Furthermore, we find that LLMs' ability in detecting intentions does not translate to their pragmatic response generation, highlighting a key distinction between merely detecting intentions and pragmatically acting on them in a generative setting.
Finally, we explored approaches to improve LLMs' pragmatic response abilities.  Using chain-of-thought prompting to make models explicitly spell out intentions before generation has minimal effects in addressing these limitations. While providing the oracle true intentions yielded better performance, models still significantly struggle to effectively utilize these intentions in response generation. 

Overall, our findings indicate a significant gap in current LLMs' ability in pragmatic understanding. This emphasizes the need for better mechanisms to infer communicative intentions \textit{and} their effective usage, to enhance pragmatic communication.

%% file: sections/3-response_generation.tex
\section{Pragmatic Response Generation}

We introduce a new framework to evaluate pragmatic generative ability of models---to understand and infer \textit{implicit} intentions, and \textit{use} it to generate pragmatic responses to non-literal utterances. 
 
\paragraph{Setup} 
Our evaluation setup (pictured in Figure \ref{fig:figure1}) measures LLMs' pragmatic response generation by comparing it to reference dialog chains under the intended true meaning and under a literal misinterpretation. 
Specifically, it requires:
\begin{itemize}[leftmargin=1.25em,itemsep=0.3em,]
    \item \textbf{Context $C$}: A short narrative involving 2 or more characters.
    \item \textbf{Non-literal Utterance $U_1^N$}: A speaker-generated utterance using non-literal language.
    \item \textbf{True Intention $I_T$}: The actual intended meaning of the speaker. 
    \item \textbf{Incorrect Literal Intention $I_L$}: An incorrect literal interpretation of the speaker's intention.
    \item \textbf{Reference Dialog Chains based on $I_T$ and $I_L$}: Speaker alternatively uses direct language to convey intentions $I_T$ as $U_1^T$ and $I_L$ as $U_1^L$. The listener responds accordingly to $U_1^T$ and $U_1^L$, with $U_2^T$ and $U_2^L$ respectively. See Figure \ref{fig:figure1}. 
    
\end{itemize}

\paragraph{Evaluating Pragmatic Understanding}
\label{secc:eval_framework}
Our framework evaluates the extent to which LLMs' generated responses reflect an understanding of the underlying speaker's intention. We operationalize this into an automatic metric by using similarity measurements. Ideally, if LLMs can accurately infer and use the intent to generate \textit{cooperative} responses using direct language, they should respond as if the non-literal utterance was instead communicated literally. Thus, if an LLM generates pragmatic cooperative responses, the response should be closer in similarity to response generated under the true intention than to one based on the literal interpretation i.e., the relation $ sim(U_2^N, U_2^T)$ $>$ $sim(U_2^N, U_2^L)$ should hold under the context $C$. 

\begin{figure*}[th]
    \centering
    \includegraphics[trim={0 1em 0 1em}, width=\textwidth]{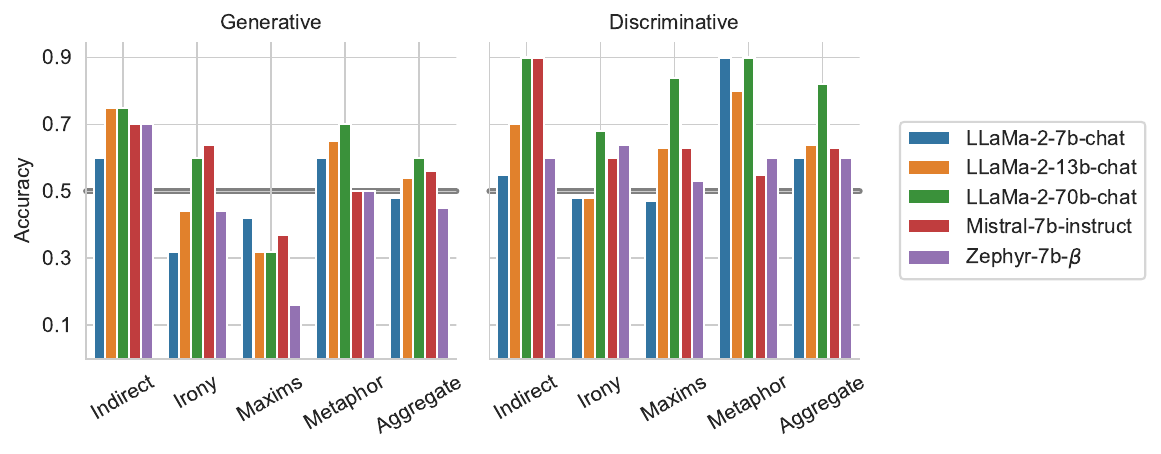}
    
    \caption{Comparison between intention resolution in response generation vs intention detection by LLMs. On average, LLMs fine the generative setting harder than the discriminative setting for non-literal language use.}
    \label{fig:gen_vs_disc}
\end{figure*}

\paragraph{Data} 
\label{secc:response:data}

\citet{hu2022fine} evaluate intention detection with a context $C$, a single non-literal utterance $U_1^N$, and verbalized intents that include a literal intent $I_L$ and true intent $I_T$. To instantiate our framework, we augment this data with dialog chains $(U_1^L, U_2^L)$ conditioned on the literal intent $I_L$ and $(U_1^T, U_2^T)$ conditioned on the true intent $I_T$. We use GPT-4 to get reference chains (See Appendix \ref{app:gold_dialog_chains}). 

We consider four non-literal language phenomena from \citet{hu2022fine}:\footnote{\citet{hu2022fine} have other tasks but we do not include them (e.g., Deceits is too non-cooperative).}
\begin{enumerate}[leftmargin=1em,itemsep=0.3em]
    \itemsep0em
    \item \textsc{Indirect Speech.} Speakers phrase requests indirectly, such as questions (``Can you pass the salt?'') or statements (``It is cold in here'').
    \item  \textsc{Irony.} Speakers use irony to mean the opposite of what they say.  Irony is not explicitly defined in the context $C$, but $C$ may include information about characters' emotional states.
    \item \textsc{Maxims of conversation.} In this task, speakers flout one of Grice's maxims. 
    \item \textsc{Metaphor.} In this task, the speaker uses metaphors to draw comparisons between entities in a non-literal sense. 
\end{enumerate}

\paragraph{Models}
\label{secc:response:models}
We evaluate five state-of-the-art LLMs: Llama2-7B-chat, Llama2-13B-chat, Llama2-70B-chat, Mistral-7B-Instruct-v0.2 and Zephyr-7B-$\beta$ instruction finetuned models. We generate candidate listener responses $U_2^N$ using these models, given the preceding context $C$ and the speaker's non-literal utterance $U_1^N$. We exclude closed-source API models (GPT-3.5/4/variants) from our evaluation suite, since we follow \cite{hu2022fine}'s discriminative setup which requires access to models' input token probabilities. Please refer to Appendix \ref{app:response_models} for generation details.  

\paragraph{Evaluators} 
\label{secc:response:evaluators} 
\subparagraph{Human Evaluation} Since LLM responses are intended for human conversational partners, we solicit human judgments to check \emph{whether understanding of the true intent is reflected in the generated response}. We employ 9 students from our institution to evaluate whether Mistral-Instruct responses successfully capture the true intended intention $I_T$ behind the speaker's non-literal utterance $U_1^N$, within the given context $C$. We choose Mistral-Instruct arbitrarily, since it is reported to surpass Llama-2-13B-chat model \cite{jiang2023mistral} and is similar in performance to Llama-2-70B-chat \cite{zheng2023judging}. We find that our annotators have a good agreement.\footnote{pairwise agreement = $0.8$, Krippendorff's $\alpha = 0.6$}

\subparagraph{GPT-4 Contextual Similarity} Separately, we tasked \texttt{GPT-4} with a \emph{contextual similarity evaluation} (cf. Section \ref{secc:eval_framework}): Given the context $C$, the speaker's true intended meaning $I_T$, and the Mistral-Instruct generated response $U_2^N$, \emph{\texttt{GPT-4} uses all the information} to identify whether $U_2^N$ is more similar to the reference response conveying the true intention ($U_2^T$) or the one with the incorrect literal intention ($U_2^L$). We find that  \texttt{GPT-4} agrees well with human annotators.\footnote{We average across individual pairwise agreements of each annotator with \texttt{GPT-4} (pairwise agreement = $0.77$, $\sigma=0.05$; Krippendorff's $\alpha=0.54$, $\sigma=0.1$)} 

\subparagraph{Non-Contextual Embedding Similarity with Llama-3-8B-Instruct} We also measure the non-contextual cosine similarity of $U_2^N$ embeddings with reference response conveying the true intention ($U_2^T$) versus the incorrect literal intention ($U_2^L$). Using LLM2Vec \cite{llm2vec}, we obtain text embeddings from Llama-3-8B-Instruct. The similarity measured using Llama-3 embeddings generally aligns with human annotations, though it agrees less than GPT-4's contextual similarity evaluation.\footnote{Similar to GPT-4, we average across individual pairwise agreements of each annotator with \texttt{Llama-3-embeddings} (pairwise agreement = $0.74$, $\sigma=0.005$; Krippendorff's $\alpha=0.46$, $\sigma=0.01$)} Additionally, we experiment with contextual embedding similarity variations \cite{yerukola2023don}, where the context $C'$ can be $I_T$, $I_L$, or turn-1 responses $U_1^T$ or $U_1^L$. However, this setting performed worse. We hypothesize that non-literal language nuances are harder to be captured by embeddings alone.

Thus, we use the better performing \texttt{GPT-4} contextual similarity evaluation as a proxy for our evaluation paradigm in all our subsequent experiments.

%% file: sections/4-experiments_for_response.tex
\section{Results on Pragmatic Response Generation}
\label{sec:response_experiments}
\label{sec:results}
In this section, we analyze how well LLMs can generate contextually relevant responses. We compare our proposed generative approach, which evaluates implicit understanding in responses to $U_1^N$, against a discriminative multiple-choice setup as in \citet{hu2022fine}, which evaluates intention detection in $U_1^N$ utterances.


\paragraph{Results}
\label{ssec:response:results}
Figure \ref{fig:gen_vs_disc} indicates that LLMs exhibit better performance in responding to \textsc{Indirect Speech} among various non-literal language types, potentially due to conventionalization of  responses, or explicit descriptions of requests completed seen during training \cite{hu2022fine}. Models perform the worst at responding to flouted \textsc{Maxims}, performing worse than chance. For instance, models fail to detect the attempt to change the subject in ``Oh, it's such a pleasant day today'' amidst a discussion about a ``bad date''. Llama-2 models exhibit marginally better metaphorical language understanding (\textsc{Metaphors}) compared to Mistral and Zephyr models. In the Llama-2 family, we see that models perform better with increasing size. 
In aggregate, we see that LLMs perform at or near chance in generating an appropriate response that reflects having inferred the true intent. 

\paragraph{Comparison against Discriminative Intention Detection}
We follow the multiple-choice setup as in \citet{hu2022fine} (details in Appendix \ref{app:jen_disc_setup}). In Figure \ref{fig:gen_vs_disc}, we consistently see that models find it easier to detect true intentions in social situations that involve flouting conversational norms (\textsc{Maxims}) in a multiple-choice setup. However, they struggle with \textit{using} this potentially inferred understanding in pragmatic response generation. 

We see that trends do not remain consistent across different models and phenomena, and that on average, models struggle more in the generative setting. We hypothesise that in a discriminative setup, the model can access all options, thus it knows the answer form in advance and has the ability to evaluate the answers contrastively. However, in a generative setup, the model's generation is free-form, requiring consistency and minimal compounding errors. This underscores the importance of evaluating model performance in \textit{both} discriminative and generative settings to obtain a better understanding of LLMs' pragmatic understanding.

%% file: sections/7-expanding.tex
\section{Chain-of-Thought Prompting for Pragmatic Response Generation}
Motivated by the ability of LLMs to detect intentions in some phenomena, we explore ways to improve their understanding of implicit intentions and, thereby enhancing their capability to generate pragmatic responses using chain-of-thought prompting (CoT) \cite{camburu2018snli, wei2022chain}.

\paragraph{Experiments using Chain-of-Thought}
\label{ssec:cot}
In our experiments with CoT, we first generate an inferred intention and then a response (unless otherwise specified). We examine how response generation performance is affected by introducing varying levels of oracle cues at the inferred intention generation step, organized by increasing amounts of ``hand-holding'': 
\begin{enumerate}[label=(\arabic*),topsep=0.5em,start=0,itemsep=0em]
    \item No oracle information (Naive)
    \item Counterfactual reasoning to clarify the non-literal utterances (no inferred intention here)
    \item Questioning a specific phenomenon (e.g., 'is Kelly being ironic')
    \item Merely indicating non-literal language use
    \item Identifying the phenomenon (e.g., 'Kelly is being ironic')
    \item Providing the true intention as CoT (no model-generated inferred intention here)
    \item Providing true intention \textit{and} phenomenon information (e.g., ``Kelly wants to read a lot and is using irony to convey it'')
\end{enumerate}


\begin{figure}
    \centering
    \includegraphics[trim={0 1em 0 0}, width=\columnwidth]{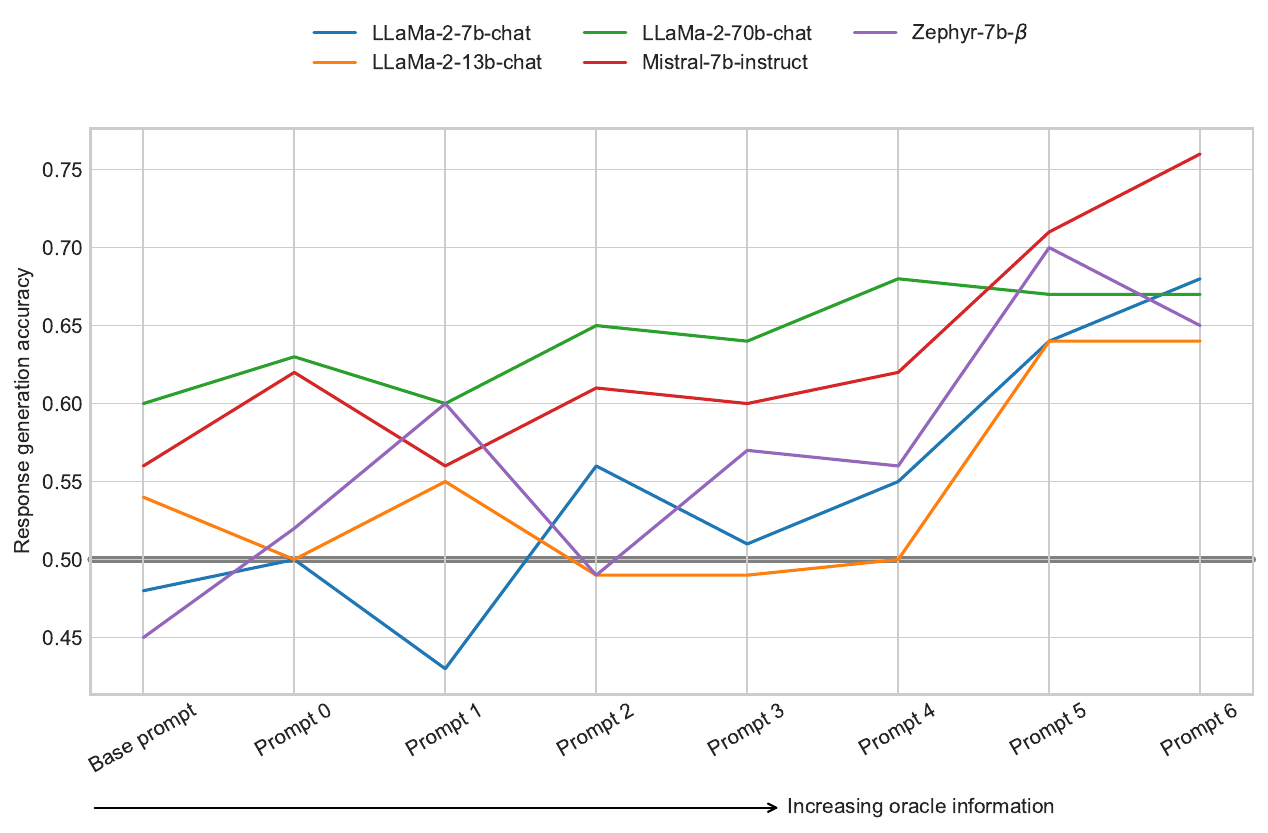}
    \caption{Results from experiments with CoT prompting show that performance is highest when providing oracle true intention, and lowest with no oracle information.}
    \label{fig:cot_experiments}
\end{figure}

\paragraph{Results}
Figure~\ref{fig:cot_experiments} illustrates that specifying the type of non-literal language used along with the speaker's true intent (\texttt{Prompt 6}) significantly improves the model's ability to generate appropriate responses, with top-performing Mistral-Instruct achieving ~75\% accuracy. Even providing subsets of this, such as just the true intention (\texttt{Prompt 5}), generally improves performance. In these cases, the task essentially becomes leveraging the provided oracle true intention in response generation. However, despite this simplification, there is still room for significant improvement in pragmatic response generation.

Intuitively, if models can accurately infer these intention cues themselves, they could generate pragmatic responses. We observe ta slight improvement in performance (on average) when no oracle information is provided (\texttt{Prompt 0}) or when prompted for counterfactual reasoning regarding the non-literal expression (\texttt{Prompt 1}). Providing explicit cues about the phenomenon (e.g.,`Kelly is being ironic' vs.\ `Is Kelly being ironic?') help slightly (\texttt{Prompts 2-4}), although not as significantly as providing the true intention.

These findings highlight the importance of explicitly modeling intention in LLMs, indicating that response accuracy to non-literal language can improve with such approaches. Overall, there is a clear need for: (a) better learning mechanisms to help models effectively disentangle the linguistic strategies used and communicative intent (e.g., recognizing how exaggeration can create irony to highlight disagreement), and (b) effective utilization of learned intentions during response generation.

%% file: sections/2-relatedwork.tex
\section{Related Work}
\paragraph{Non-literal language understanding in LLMs}
Recent work has proposed several ways to evaluate LLMs' ability to interpret non-literal language, including implicature \cite{ruis2023goldilocks,kim2023is}, figurative language use \cite{liu-etal-2022-testing,chakrabarty2022flute,gu2022just,chakrabarty2022s, wachowiak-gromann-2023-gpt, lai2024survey}, detecting profundity \cite{herrera-berg-etal-2023-large}, broader benchmarks for social language understanding \cite{choi-etal-2023-llms} and various pragmatic phenomena \cite{li2017dailydialog,zheng2021grice, hu2022fine}. \citet{kim2023is} also find that chain-of-thought helps improve a model's ability to interpret the use of implicatures. These tasks have focused on evaluating models' ability to \textit{interpret} the true intent underlying an utterance, but not \textit{respond} to it as we do in this work. 
Another line of work has considered LLMs' mentalizing abilities using false belief tasks \cite{shapira2023clever} or question answering \cite{le-etal-2019-revisiting,kim-etal-2023-fantom}. \citet{zhou2023far} consider a task that evaluates how models respond using knowledge of other agents' mental states. 

\paragraph{Generating responses based on inferred intents} Some work has presented resources for intent or emotion-conditioned response generation, where a conversational agent must respond conditioned on a particular intent or emotion. \citet{li-etal-2017-dailydialog} and \citet{rashkin-etal-2019-towards} present datasets of dialogues annotated with discrete emotion or intent labels. \citet{zhang-zhang-2019-hierarchy} and \citet{chen-etal-2022-emphi} present approaches to modeling intent explicitly. \citet{gu-etal-2022-just} generate explicit scene elaborations to improve figurative language understanding. While these works consider conditioning on intent, they do not explicitly focus on generating or evaluating responses to non-literal language use.

%% file: sections/8-conclusion.tex
\section{Summary}
We propose a new framework to evaluate how well LLMs understand intentions and respond to non-literal language, moving beyond previously employed multiple-choice settings. Our results show that LLMs often struggle to generate contextually relevant responses. While chain-of-thought prompting to spell out inferred intentions offers marginal improvements, explicitly providing oracle intentions and cues, such as for irony, significantly enhances performance. These findings highlight the current limitations of LLMs in pragmatic understanding, suggesting that improved learning mechanisms to explicitly model intentions and linguistic strategies could significantly enhance conversational abilities.

%% file: sections/9-limitations.tex
\section{Limitations \& Ethical Considerations}
Despite taking the first step towards proposing a new generative framework for evaluating intention resolution in LLMs, there are several limitations and ethical concerns, which we list below.

\paragraph{Limited Context Scope} In this study, our primary focus is the evaluation of intention understanding and using it in pragmatic response generation. Future work should explore introducing other forms of context into the pragmatic generation pipeline, such as richer social and power dynamics \cite{antoniak2023riveter}, emotional states \cite{zhou2023cobra}, and external knowledge \cite{ghazvininejad2018knowledge}, all of which can significantly contribute to varied levels of pragmatic understanding.

\paragraph{Amount of context} In our experiments, we opted to include short 1-3 sentence stories. Future work can explore longer stories and include more preceding dialog turns. We hypothesize that more context will make this task more challenging, and we would need nuanced ways of understanding intentions at different turns.

\paragraph{Limited number of non-literal phenomenon} We explore the evaluation of only four phenomena: \textsc{Indirect Speech}, \textsc{Irony}, \textsc{Maxims}, and \textsc{Metaphors}. Future work should consider other types of figurative language, such as cultural metaphors \cite{kabra2023multi}, visual metaphors \cite{liu2022testing}, idioms, proverbs, etc. Expanding the scope to include these elements would provide a more comprehensive understanding of LLMs' capabilities in interpreting nuanced language.

\paragraph{Potentially Inconsistent Human Evaluation} In our work, we employ only 9 expert human annotators and assume human judgments as the gold standard. Concurrent work has shown that human evaluation might not always be consistent \cite{clark2021all, karpinska2021perils}; however human judgments continue to be the gold standard for evaluating open-ended text generation. 

\paragraph{Potential effects on Factuality} In our work, we show that LLMs struggle with responding pragmatically to non-literal language. Training approaches which might help with better intention modeling to handle non-literal language may potentially affect faithfulness or factuality of LLMs responses.

%% file: sections/10-acknowledgement.tex
\section{Acknowledgements}
We would like to thank our student annotators for helping us with intention resolution annotations. We thank OpenAI for providing researcher credits to access GPT-4. This project is funded in part by DSO National Laboratories and an Amazon Research Award, Spring 2023 CFP. Any opinions, findings, and conclusions or recommendations expressed in this material are those of the author(s) and do not reflect the views of Amazon.

%% file: sections/appendix.tex
\section{Pragmatic Response Generation}
\subsection{Data}
We consider four non-literal language phenomenon from \citet{hu2022fine}:
\begin{enumerate}[leftmargin=1.25em,itemsep=0em,topsep=0em]
    \itemsep0em
    \item \textsc{Indirect Speech} - 20 examples
    \item  \textsc{Irony} - 25 examples
    \item \textsc{Maxims of conversation} 20 examples
    \item \textsc{Metaphor} - 20 examples
\end{enumerate}
These examples were were manually curated by expert researchers to cover a broad
range of non-literal phenomena and elicit individual differences among humans. 

\subsection{Gold Dialog Chains based on $I_T$ and $I_L$}
\label{app:gold_dialog_chains}
In our experiments, we use \texttt{GPT-4} to generate counterfactual gold dialog chains based on $I_T$ and $I_L$. We first generate $U_1^L$ by providing \texttt{GPT-4} with only the incorrect literal intention $I_L$ to prevent any possible contextual inference from $C$ from being reflected in $U_1^L$. Next, we generate $U_1^T$ by providing \texttt{GPT-4} with the context $C$ and true intention $I_T$, mimicking an ideal direct cooperative utterance generation setting. We follow the same setting to generate $U_2^L$ and $U_2^T$ from $U_1^L$ and $U_1^T$, respectively. 

\subsection{Models}
\label{app:response_models}
We use the following prompt to generate responses using LLMs:
\begin{verbatim}
Generate a short, concise single sentence 
response.
Context: {context}
{speaker's non-literal utterance}

Generate a co-operative response without 
any non-literal language as 
{listener character name}. 

{listener character name} replies, 
\end{verbatim}
We use temperature sampling with temperatures of $0.3$ and $0.5$ to generate responses, averaging the results from both settings. We find that a lower temperature$=0$ results in incoherent, uninteresting responses, while a temperature $>0.5$ leads to very verbose responses that digress from the main context and extrapolate excessively.

\subsection{Evaluators}
\label{app:gpt4_eval}
For \texttt{GPT-4} as an evaluator, we provide the following prompt: 
\begin{verbatim}
Task: You will read a short story. The story 
will be followed by a question. Your 
task is to decide which response option is 
closest to the 'Generated Response'. 
The answer options are 1 or 2.

Scenario: {Context}
{speaker's non-literal utterance U_1^N}
Intention: {True intention}

Generated Response:  
{Person2} replies, {U_2^N from model}	 
 		 
Compare the below utterances to the 
Generated Response. Which of the 
below utterances is closest to the 
above Generated Response? 
Options: 
1. {U_2^T from true intention dialog chain}
2. {U_2^L from incorrect literal intention 
dialog chain}

Answer (option number 1 or 2): {number}
\end{verbatim}
We randomly shuffle intentions provided as options 1 and 2. We set the temperature=$0$ to ensure the stability of the evaluation. 

For human evaluation, we provide the following prompt:
\begin{verbatim}
Task: You will be provided a short story,
an utterance by one of the characters 
in the story (person1). Person1 uses 
non-literal language (like irony). 
Person2 from the story responds to  person1's 
utterance. The task is to identify if 
the "true intention" (provided) is 
resolved/understood in person2's response 
or not.

Make a binary yes/no choice. 
\end{verbatim}
We employ 9 students from our institution -- 6 women, 3 men (20-30 age group) living in the United States of America. 
\section{Discriminative Setup}
\label{app:jen_disc_setup}
We follow setup in \citet{hu2022fine} for our discriminative setup comparison. They use a the multiple-choice setup. They compute the probability of answer options -- true intention $I_T$ and literal misinterpretation $I_L$ -- given the context $C$, the speaker's non-literal utterance $U_1^N$, and task instructions. We measure accuracy as assigning the highest probability to the correct answer token (e.g., ``1'', ``2''). 
We follow the same prompt template as \citet{hu2022fine}:
\begin{verbatim}
Task: You will read short stories that 
describe everyday situations. Each 
story will be followed by a multiple-
choice question. Read each story and 
choose the best answer. Your task is 
to decide what the character in the 
story is trying to convey. The answer 
options are 1 or 2.

Scenario: {context} {dialog}. 
What might {person1} be trying to convey?
Options:
1) {option1}
2) {option2}
Answer: 
\end{verbatim}

\section{Chain-of-thought Prompting}
\label{app:cot}
Please refer to \label{fig:cot_templates} for the chain-of-thought prompting templates used for all the models

\subsection{Inferred Intention vs Response Accuracy}
We evaluate similarity of CoT generated intents with the true intent and the incorrect literal intent using \texttt{GPT-4}. We follow a similar prompt as \texttt{GPT-4} evaluator in Appendix \ref{app:gpt4_eval}.  We observe in Figure \ref{fig:inferred_intention_response_acc} that a model that is able to correctly infer the underlying true intention is also  better at generating contextually relevant responses, corroborating our finding from \textsc{Prompt 5-6} in Section \ref{ssec:cot}.

\begin{figure}
    \centering
    \includegraphics[width=\columnwidth]{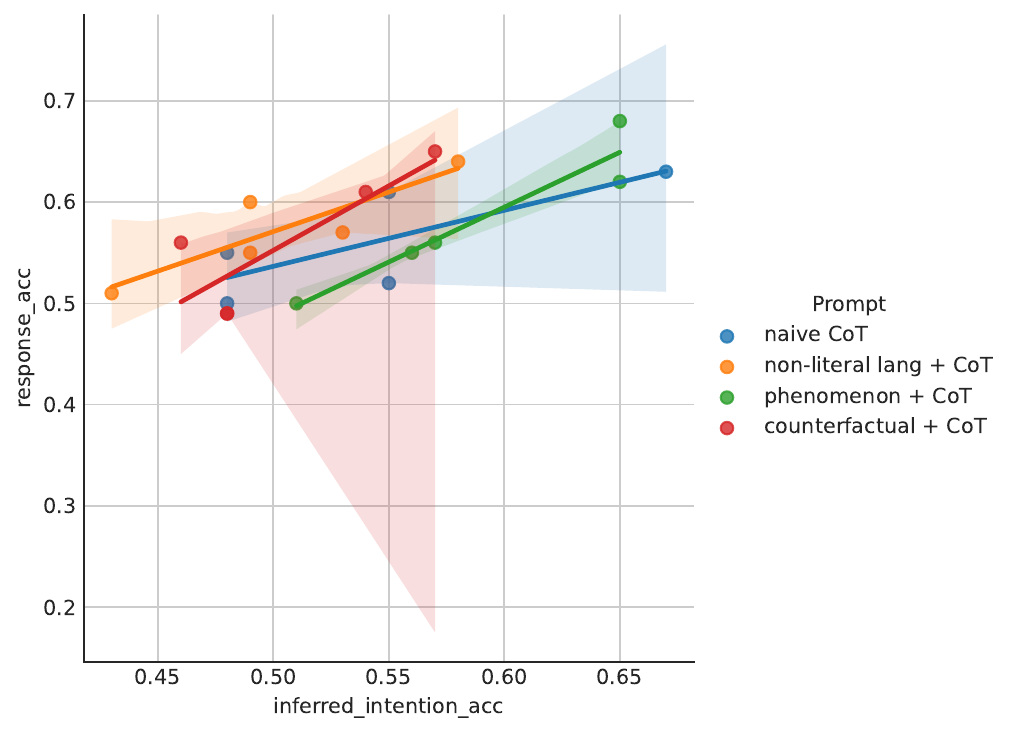}
    \caption{Positive correlation between inferred intention accuracy and pragmatic response accuracy.}
    \label{fig:inferred_intention_response_acc}
\end{figure}

\begin{figure*}[!t]
    \centering
    \includegraphics[width=2.2\columnwidth]{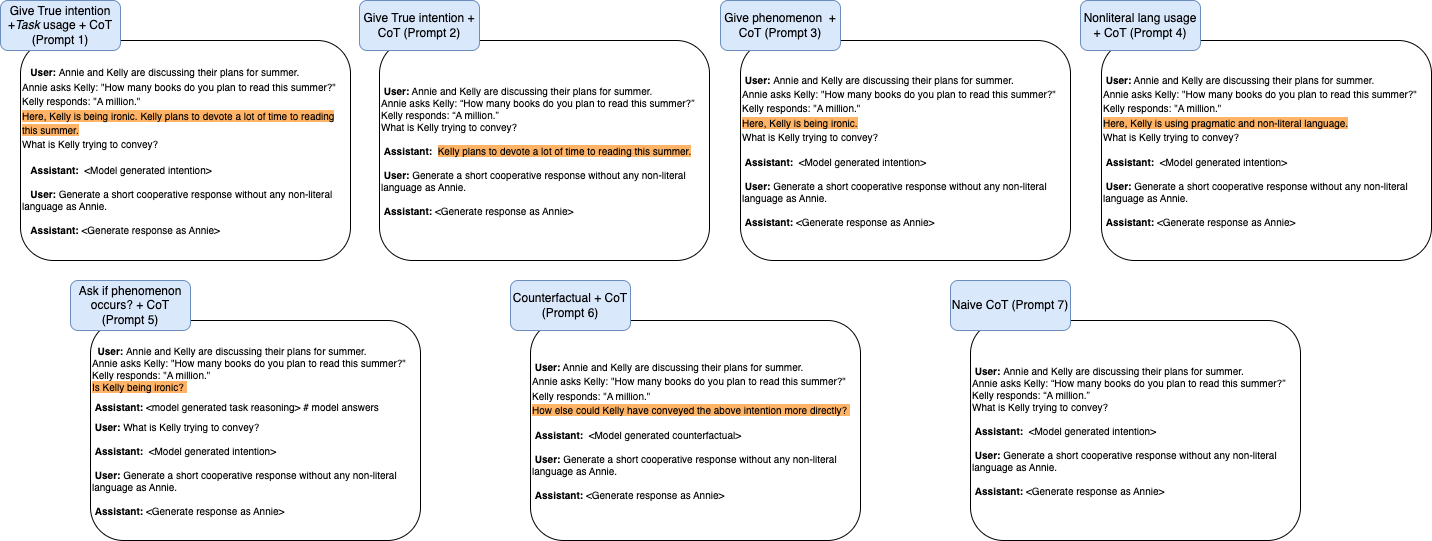}
   \caption{Chain-of-thought Prompting templates used in Section \ref{ssec:cot}. Orange highlighted text is the explicitly provided oracle information. }
    \label{fig:cot_templates}
\end{figure*}